\PassOptionsToPackage{table}{xcolor}
\documentclass[dvipsnames,sigconf=true, nonacm=true, review=false, anonymous = false]{acmart}
\usepackage{booktabs} 
\usepackage{layouts}

\copyrightyear{2024}
\acmYear{2024}
\setcopyright{rightsretained}
\acmConference[GECCO '24]{Genetic and Evolutionary Computation Conference}{July 14--18, 2024}{Melbourne, VIC, Australia}
\acmBooktitle{Genetic and Evolutionary Computation Conference (GECCO '24), July 14--18, 2024, Melbourne, VIC, Australia}
\acmDOI{10.1145/3638529.3654100}
\acmISBN{979-8-4007-0494-9/24/07}

\usepackage{amsmath}
\usepackage{xspace}
\usepackage{xcolor}
\usepackage{dsfont}
\usepackage{rotating}
\usepackage{makecell}
\usepackage{multirow}

\usepackage{amsthm}
\usepackage{amsmath}
\renewcommand{\epsilon}{\varepsilon}

\newcommand{\swpkg}[1]{\textsc{#1}}    
\newcommand{\feature}[1]{\texttt{#1}}  
\newcommand{\reffig}[1]{Fig.~\ref{#1}} 
\newcommand{\reftab}[1]{Tab.~\ref{#1}} 


\begin{document}
\title[Impact of Training Instance Selection on AAS Models for Numerical Black-box Optimization]{Impact of Training Instance Selection on Automated Algorithm Selection Models for Numerical Black-box Optimization}

\author{Konstantin Dietrich}
\affiliation{
  \institution{Technische Universität Dresden \& ScaDS.AI}
  \city{Dresden}
  \country{Germany}
}
\email{konstantin.dietrich@tu-dresden.de}

\author{Diederick Vermetten}
\affiliation{
  \institution{Leiden Institute for Advanced Computer Science}
  \city{Leiden}
  \country{The Netherlands}
}
\email{d.l.vermetten@liacs.leidenuniv.nl}

\author{Carola Doerr}
\orcid{0000-0002-4981-3227}
\affiliation{
  \institution{Sorbonne Universit\'e, CNRS, LIP6}
  \city{Paris}
  \country{France}}
\email{carola.doerr@lip6.fr}

\author{Pascal Kerschke}
\affiliation{
  \institution{Technische Universität Dresden \& ScaDS.AI}
  \city{Dresden}
  \country{Germany}
}
\email{pascal.kerschke@tu-dresden.de}

\begin{abstract}
The recently proposed MA-BBOB function generator provides a way to create numerical black-box benchmark problems based on the well-established BBOB suite. Initial studies on this generator highlighted its ability to smoothly transition between the component functions, both from a low-level landscape feature perspective, as well as with regard to algorithm performance. This suggests that MA-BBOB-generated functions can be an ideal testbed for automated machine learning methods, such as automated algorithm selection (AAS).

In this paper, we generate 11\,800 functions in dimensions $d=2$ and $d=5$, respectively, and analyze the potential gains from AAS by studying performance complementarity within a set of eight algorithms. We combine this performance data with exploratory landscape features to create an AAS pipeline that we use to investigate how to efficiently select training sets within this space. We show that simply using the BBOB component functions for training yields poor test performance, while the ranking between uniformly chosen and diversity-based training sets strongly depends on the distribution of the test set. 
\end{abstract}

\maketitle

\section{Introduction}

Over the last decades, the number of algorithms which tackle continuous optimization problems via iterative search procedures has drastically increased~\cite{back2023evolutionary}. With this development, the question of selecting an appropriate algorithm for a given problem has become harder to answer~\cite{Rice76}. Not all algorithms are equally proficient in all types of optimization problems, resulting in significant complementarity between different optimizers. As such, algorithm selection techniques are becoming increasingly popular, since they have the potential to exploit some of this performance complementarity~\cite{KerschkeHNT19}. 

At their core, algorithm selectors map information about the problem to be solved to a specific algorithm instance that is expected to perform particularly well on this problem. In its simplest form, this mapping can be based on high-level information, such as the number, types, and ranges of the decision variables and information about the available budget~\cite{MeunierDRT20}. In the strict black-box setting, we cannot expect to have more information about the problem instance at hand. However, one can decide to invest a small fraction of the budget to extract properties of the problem instance, for example via so-called exploratory landscape analysis (ELA) techniques~\cite{mersmann2011exploratory}, which are designed to characterize important properties such as the ruggedness of a problem, its separability, the fit of linear or quadratic models, and so on. 

ELA-based algorithm selection methods have shown quite some promise, highlighting that using some samples to collect more detailed information about the problem is worthwhile compared to just running a single algorithm, as this allows the selector to exploit additional layers of performance complementarity~\cite{KerschkeT19,BelkhirDSS17}. However, a series of recent works indicate that the good performances that these selectors achieve within well-known benchmark suites hardly transfer to other problem collections, even when these are chosen to be very similar to the original ones~\cite{kostovska2022per,vskvorc2022transfer}. These observations raise an important question regarding the generalization ability of the algorithm selectors. 
In the broader context of benchmarking, research suggests that careful selection of problems can have a clear impact on the results of benchmarking~\cite{GECCO2022SELECTOR} and performance prediction of optimization algorithms~\cite{nikolikj2023rf}, suggesting that the same should be true within an algorithm selection context.

To assess the extent to which the choice of training instances influences performance on different test sets, we explore the recently proposed MA-BBOB problem generator~\cite{mabbob_arxiv}, which facilitates the generation of large numbers of problem instances. Precisely, we collect algorithm performance data for $11\,920$ MA-BBOB problems in dimensions $2$ and $5$, respectively. We then split these sets into training and test sets. We compare three methods for instance selection: (1) using the BBOB functions (which are used in the MA-BBOB framework as components to generate new problems), (2) random selection, and (3) a greedy selection maximizing diversity with respect to the ELA features. We observe that, while training on random instances gives the best performance on the full set of problems, the greedy selection leads to better performance on more equally distributed testsets. 

\section{The MA-BBOB Generator}
\label{sec:mabbob}

Within numerical derivative-free black-box optimization, the BBOB suite, which has been proposed as part of the COCO platform~\cite{hansen2021coco}, has become one of the most used sets of benchmarking problems. The single-objective, noiseless version of BBOB consists of 24 scalable functions, of which different instances can be created by applying a set of transformations~\cite{bbobfunctions}. These transformations preserve the high-level structure of the problem and allow for assessing the presence of certain biases of the algorithms, although it is not guaranteed that low-level features are consistent between all instances of a problem~\cite{long2023bbob, vskvorc2020understanding}. 

While the BBOB suite provides a useful setup for benchmarking optimization algorithms, it is not specifically designed to handle techniques from \textit{(per-instance) Automated Algorithm Selection (AAS)} \cite{KerschkeHNT19}. AAS methods which use BBOB as a test-suite either use a train-test split based on functions, where the distance between the training and test sets can become quite large (\textit{leave-one-problem-out} scenario) \cite{SeilerPKT22featurefree}, or based on instances, where the distance becomes very small (\textit{leave-one-instance-out} scenario),~\cite{nikolikj2023rf}. 

Several methods have been proposed in the last years to obtain functions that fill the instance space spanned by the BBOB functions, using Genetic Programming~\cite{MunozS20spacefillingBBO,long2023challenges}, random function generation~\cite{tian2020arecommender}, or recombining the BBOB functions via affine combinations~\cite{DietrichM22affinebbob}.

Since the latter approach is particularly scalable while preserving interesting problem properties, we focus on affine combinations of BBOB functions in this work. More precisely, we consider the multi-affine BBOB function generator (MA-BBOB) proposed in~\cite{MABBOBautoML}. Extending the initial  concepts by~\cite{DietrichM22affinebbob} and~\cite{ABBOB-GECCO}, the MA-BBOB generator allows to combine an arbitrary number of the 24 BBOB functions through affine combinations. This generator incorporates a rescaling of the component functions to ensure each component has a similar impact on the final function value if given equal weight. To ensure the global optimum is known and unbiased, a uniform sample inside the domain is picked as the optimum, and the component functions are translated to have their optimum at this location, with an optimal function value of $0$ to allow for easy comparison. The MA-BBOB functions are accessible through the IOHexperimenter environment~\cite{iohexperimenter}, which facilitates the data generation via the existing interface to, e.g., the large body of algorithms available in the Nevergrad platform~\cite{nevergrad}.   

\section{Experimental Setup}\label{sec:setup}

\paragraph{Function Generation}\label{sec:functions}
To explore the range of problems which can be generated using MA-BBOB, we opt not to use the default instantiation procedure proposed in~\cite{MABBOBautoML}. Instead, we cover the full range of the number of active component functions combined, from $2$ to $24$. Since the function combinations with many combined components are more likely to be similar to each other, we divide the number of functions generated into two parts. For 2 to 5 active components, we generate $2\,000$ functions each, while for 6 to 24 active components, we limit ourselves to 200 functions each. This results in a total of $11\,800$ functions per problem dimensionality. Each function has a uniformly distributed location of the optimum and uses, for each active component function, an instance that is chosen uniformly at random among the first 100 instances available in COCO's BBOB implementation. The weights used in the affine combination are also generated independently and uniformly at random for each active component, before being normalized to sum up to 1. 

In addition to the generated MA-BBOB functions, we also make use of the component functions. Specifically, this means that we use the rescaled versions of the original BBOB problems. The rescaling, proposed in~\cite{MABBOBautoML}, uses scaling factors that ensure each function has a target of $10^2$ which is similarly easy/hard to find. 
In this study, we limit ourselves to the first $5$ instances of these rescaled BBOB problems (for a total of 120 problem instances). 

\paragraph{ELA calculation and selection}
\label{sec:ela}
To calculate the ELA-features, we make use of the \swpkg{Pflacco} package~\cite{Pflacco2023}. From the set of available features, we select only those that can be calculated based on a single sampling of the function space, i.e., those that do not require adaptive sampling strategies. To extract the features, we use $500d$ points obtained from the Sobol' sequence. To increase the robustness of the feature values, this sampling is repeated 5 times for each function, and the resulting values are averaged. Using this selection procedure, we obtain 74 features from the following sets: 9 \feature{ela\_meta}-~\cite{mersmann2011exploratory}, $4$ \feature{ela\_distr}-~\cite{mersmann2011exploratory}, $10$ \feature{ela\_level}-~\cite{mersmann2011exploratory}, $9$ \feature{pca}-~\cite{Kerschke2019flacco}, $13$ \feature{limo}-~\cite{Kerschke2019flacco}, $6$ \feature{nbc}-~\cite{Kerschke2015}, $17$ \feature{disp}-~\cite{Lunacek2006} and $6$ \feature{ic}-features~\cite{Munoz2015}.

Since many of these ELA-features are quite similar and have shown high correlation when calculated on the $24$ BBOB functions~\cite{RenauDDD21evoapps}, we apply a subset selection procedure based on feature correlation, computed using all of the $23\,840$ functions. First, we rigorously remove all features which returned an infeasible value for any of the analyzed functions. This leaves us with 57 features. For these features, we then determine their pairwise Pearson correlations and we count for every feature to how many other features the current feature has a Pearson correlation larger than $0.9$. We then eliminate features in an iterative fashion, always removing the one that has the largest number of features whose pairwise Pearson correlation exceeds $0.9$. This leaves us with a total of $29$ features: $8$ \feature{ela\_meta}-, $3$ \feature{ela\_distr}-, $6$ \feature{ela\_level}-, $2$ \feature{nbc}-, $6$ \feature{disp}- and $4$ \feature{ic}-features. Their pairwise correlation for the two-dimensional problems are illustrated in the upper left part of~\reffig{fig:corr_2d}. 

To allow for better comparability of feature value ranges and, therefore, problem diversity analysis, we perform min-max normalization of all ELA features. The minimum and maximum values of each feature are determined from all feature values calculated on the $11\,920$ functions within one dimensionality.

\paragraph{Training instance selection for AAS}
To create the training sets for our AAS setting, we make use of three separate techniques: 

The first, and most commonly used one, is to simply select the training instances uniformly at random from the full function set~\cite{KerschkeHNT19}. This selection method preserves the underlying distribution of the full dataset, which can be advantageous for test performance, but might hinder generalizability to more varied testsets, as discussed in the introduction. 

The second instance selection method is aimed at ensuring high diversity in feature space. We use an iterative greedy approach, where in each step we add the function whose feature values are most different to those of the functions that are already selected, where ``different'' is measured in terms of Manhattan distance calculated on the 29-dimensional ELA feature vectors. 

The third instance selection method is specific to our MA-BBOB setting: we use the component functions, i.e., the rescaled BBOB functions, as training instances. Since we have multiple instances for each BBOB function, the sets can be constructed either using a single instance for each BBOB function (resulting in a training set of size 24) or using all 5 instances for each function (for a training set of size 120). 

We use the above-described methods to select training sets of size $s \in \{24,120\}$ (for all three methods) and $s\in\{600,1\,200,1\,800,3\,600\}$ (for the first two methods only). 

To make our results more robust, we sample multiple training sets for each method and set size. Since we will use these same sets throughout testing scenarios as well, we ensure that different repetitions of the same sampling are fully disjoint. This limits the number of repetitions of the larger training set sizes. Specifically, we use $r=(10,10,5,3,3,2)$ repetitions for $s=(24,120,600,1\,200,1\,800,3\,600)$.

Note that this process is performed separately for the two different problem dimensionalities. While some studies prefer to create a single model where problem dimensionality is an additional input~\cite{SeilerPKT22featurefree}, our focus is on per-dimension models to avoid variations in ELA-features caused by dimensionality~\cite{MorganG14}.

Finally, it should be noted that another extensive instance selection procedure has been proposed recently~\cite{GECCO2022SELECTOR}. We do not incorporate this methodology, as the graph-based approach does not scale very well to the large number of instances that we consider in this work and since our scope is different in that we aim at measuring the interaction between training and test instances. 

\paragraph{Performance Data Collection}\label{sec:algs} 
Our algorithm portfolio consists of 8 optimizers, of which 6 are taken from the Nevergrad package~\cite{nevergrad}: Differential Evolution, RCobyla, ConfiguredPSO, GOMEA, DiagonalCMA, and MultiBFGS. In addition, we make use of the default configurations from two modular optimization algorithms: modDE~\cite{modde} and modCMA~\cite{modular-CMAES}. Each algorithm is run on all $11\,920$ functions described above. 
This experiment is performed in both dimensionalities, 2 and 5. For each function, we collect performance data (using the \swpkg{IOHexperimenter} package~\cite{iohexperimenter}) of $15$ independent runs ($50$ for the rescaled BBOB functions), with a budget of $2\,000 d$ function evaluations per run. We then use the trajectories of these runs to compute the area over the convergence curve (AOCC) as an anytime performance measure~\cite{eafecdf}. To be consistent with existing work on BBOB, we use a logarithmic scaling between precision values $10^2$ and $10^{-8}$ in the AOCC calculation. With this choice, the AOCC values are equivalent to the area-under-the-ECDF measure commonly used in benchmarking of numerical black-box optimization algorithms~\cite{hansen2021coco,TBB20benchmarking}.

\paragraph{Reproducibility} 
To ensure the reproducibility of our results, our data collection and feature computation setup, as well as the exact settings used to generate the $11\,920$ functions, are made available on our Zenodo repository~\cite{reproducibility_and_figures}. 
In addition to these scripts, it also contains the processed performance data from all algorithms and all code used to create, analyze and visualize the results of the AAS methods. Finally, we have also uploaded additional figures, including alternative versions of the figures shown throughout the remainder of this paper, to Figshare~\cite{reproducibility_and_figures}.

\begin{figure*}[t]
  \centering
  \includegraphics{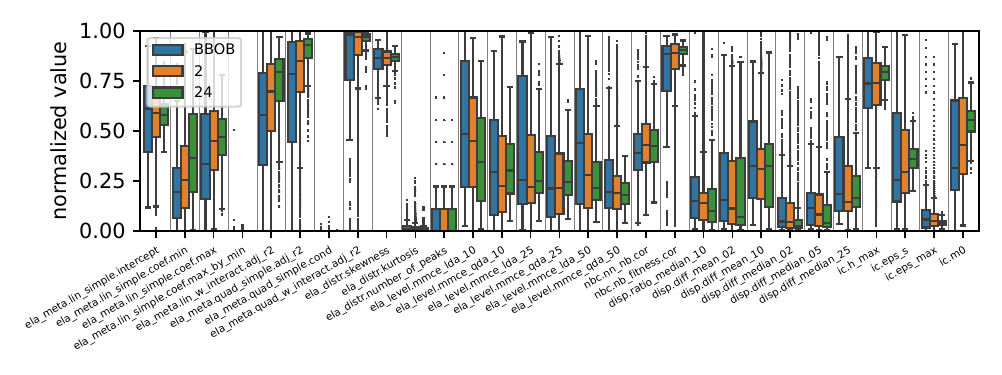}
    \caption{Boxplots of the selected ELA feature values of the two-dimensional BBOB function set (blue), and the two-dimensional function subsets with 2 (orange) and 24 (green) active component functions, respectively. Generally, the value ranges of the subgroups do not appear to be extremely different from each other. Instances with 24 active component functions become very similar which is resembled by the narrowed feature value ranges.}
     \label{fig:boxplot_2d}
\end{figure*}
\section{Results}

\subsection{Complementarity analysis}
We assess the complementarity of the MA-BBOB functions and the rescaled BBOB functions in terms of their problem properties and with respect to the performance achieved by the different solvers.

\paragraph{Problem properties}
For the complementarity analysis of the problem characteristics, we make use of the 29 ELA features selected in Section~\ref{sec:ela}. In \reffig{fig:boxplot_2d} we plot the feature values of all five instances of the two-dimensional BBOB functions next to the feature values of all generated two-dimensional problems with $2$ and $24$ active component functions. Even though we only show the two-dimensional data here (for reasons of space), the five-dimensional data goes in accord with the following observations.
As already mentioned, we expect generated problems with a larger number of active component functions to be more similar to each other. This is confirmed by \reffig{fig:boxplot_2d} in the sense that many of the feature value ranges of the problems with $24$ active components are significantly smaller than the corresponding feature value ranges of the BBOB problems or the functions with $2$ active component problems. The \feature{ic.m0} feature on the very right of \reffig{fig:boxplot_2d} gives a good example of this observation. The same can also be stated for the $2$ active component functions when comparing them to the BBOB functions but in a much less pronounced fashion.

\begin{figure*}
 \begin{minipage}{.49\textwidth}
 \centering
  \includegraphics[scale=1, trim = 0pt 70pt 0pt 90pt, clip]{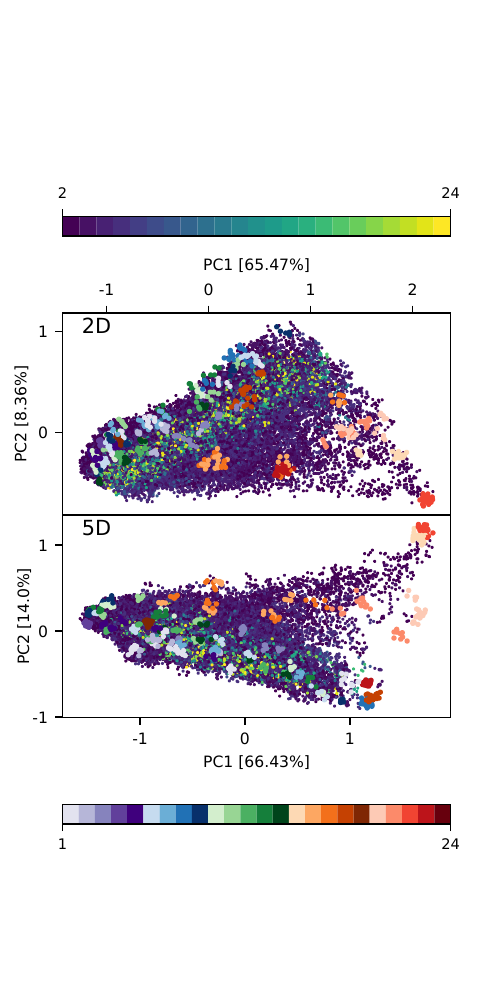}
  \caption{PCA projection of the ELA feature vectors of the 11\,800 MA-BBOB functions (small dots) and the 120 BBOB instances (larger dots),  for 2$d$ (top) and 5$d$ (bottom). Projections are fitted using the BBOB functions. The colorbar on the bottom corresponds to the BBOB function ID, whereas the one on the top is used to display how many functions were combined to create the respective MA-BBOB function. The MA-BBOB functions fill the feature space spanned by the BBOB functions but mostly remain within its convex hull.}
  \label{fig:pca_all}
 \end{minipage}
 \hfill
 \begin{minipage}{.49\textwidth}
 \centering
 \includegraphics[scale=1, trim = 0pt 0pt 0pt 35pt, clip]{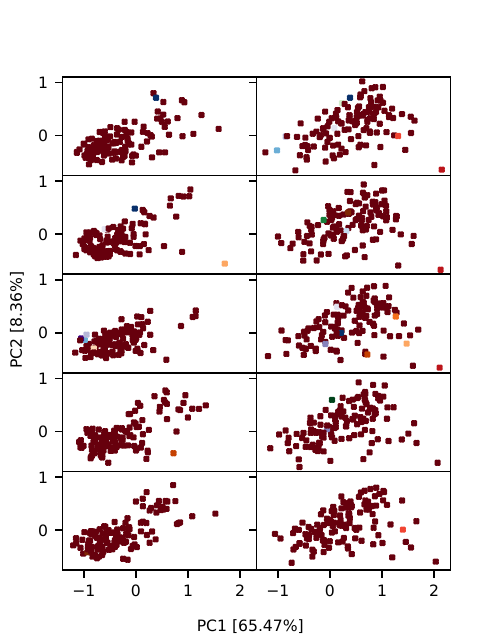}
  \caption{
  Visualization of the selected two-dimensional instance sets, using the same PCA projection as in \reffig{fig:pca_all}. The left column shows five of the uniformly sampled sets of size 120 and the right column shows five of the greedily sampled sets. Colored points are BBOB functions, using the colorbar from the bottom of  \reffig{fig:pca_all}. As expected, the greedily sampled function sets are more evenly spread in feature space while the uniform randomly sampled sets are more likely to occupy the area with the highest function density. 
  }
  \label{fig:distr_greedy_2d}
 \end{minipage}
\end{figure*}

Overall, when comparing the generated function subsets to the BBOB subset, we see that none of the feature ranges are located on entirely different ends of the scale, suggesting that the composed MA-BBOB functions are not too different from their base functions. This interpretation is further supported by \reffig{fig:pca_all}.
Here we make use of a principal component analysis (PCA) to project the 29-dimensional ELA feature vectors of all functions to 2d. 
The transformation matrices of the PCA projections are determined based on the normalized ELA vectors of the first five instances of the two- and five-dimensional BBOB problems individually and are used to project the generated functions into the so-constructed two-dimensional feature space. We color-code all instances of the BBOB functions according to the color bar on the bottom. 
The color of the generated functions is chosen according to the number of active component functions -- see color bar on top of \reffig{fig:pca_all}. 

The last observation for \reffig{fig:pca_all} is that for both analyzed dimensions, the generated functions almost exclusively remain within the convex hull spanned by the BBOB functions. 
This confirms the finding from~\cite{DietrichM22affinebbob} and expands it by showing the inability of larger numbers of component functions to traverse past these borders.
We see that functions with a two-digit active component function number reside along a diagonal in the feature space, suggesting that more diversity can be expected from single-digit active component function numbers.
Nevertheless, the generated functions complement the BBOB set in that they fill large amounts of the feature space that was left unoccupied by the BBOB functions in both dimensions.

\paragraph{Solver performance}

AAS methods aim to exploit the fact that different optimization algorithms have their own strengths and weaknesses, and no single algorithm will perform best over all possible optimization problems. However, for problem sets of limited size or diversity, it might still be the case that the potential improvements from algorithm selection are limited by a single algorithm performing exceedingly well, or from all algorithms achieving similar levels of performance on each problem. 

\begin{figure}[t]
  \centering
  \includegraphics{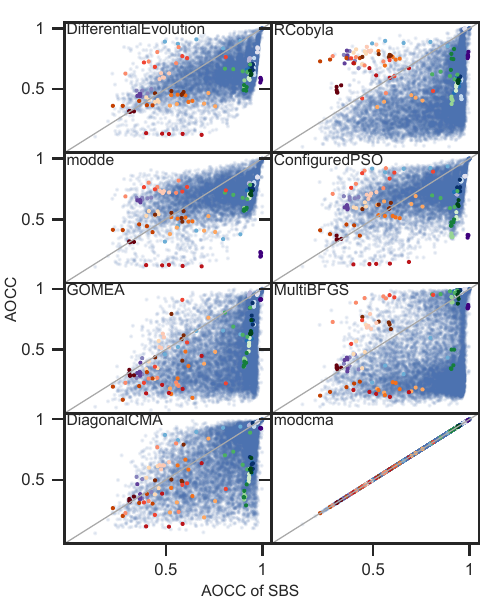}
  \caption{AOCC of the SBS on the $x$-axis vs. AOCC of the other algorithms on the $y$-axis. Each blue dot corresponds to one of the 11\,800 functions in 2$d$. BBOB functions are colored according to the colorbar in \reffig{fig:pca_all}. All points above the diagonals correspond to instances for which the respective solver beats the SBS. The fraction of instances on which at least one algorithm outperforms the SBS is 0.46. 
  }
  \label{fig:perf_compl_2d_scatter}
\end{figure}

To measure the amount of complementarity between the algorithms described in Section~\ref{sec:algs}, we make use of the difference between the algorithm which performs best on average, the \textit{Single Best Solver (SBS)}, and the average performance of the best algorithm on each instance, the \textit{Virtual Best Solver (VBS)}. The difference in their performance indicates how much can be gained by going from no algorithm selection at all to a perfect selection method which selects the ideal solver on each instance. In \reffig{fig:perf_compl_2d_scatter}, we show the relation in performance between the SBS (which, in this case, is the modcma) and each individual algorithm. Dots above the diagonal correspond to instances for which selecting the shown algorithm would lead to an improvement over the SBS. We notice that all algorithms have some functions on which they are preferable over SBS, although for COBYLA this amount is low at only 395 ($\approx3.3\%$) of the many-affine functions, even though it outperforms the SBS on 59 ($\approx50\%$) of the BBOB problems. 

\begin{figure}[h]
  \centering
  \includegraphics[width=.7\linewidth]{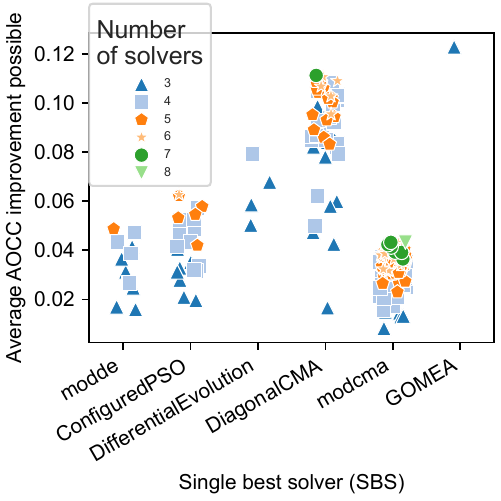}
    \caption{Potential of algorithm selection to improve over the SBS, measured in average AOCC improvement of the VBS over the SBS, for all portfolios using at least 3 of the 8 algorithms for which we have collected performance data. Data points are grouped by SBS (columns) and by portfolio size (symbol). The modCMA algorithm is very dominant in our portfolio, leaving little room for algorithm selection. When removed from the portfolio, the average VBS-SBS gap is 0.111, the second largest value obtained across subsets. 
    }
    \label{fig:perf_compl_2d}
\end{figure}

The VBS-SBS gap (computed as the difference in average AOCC of the respective methods) of the algorithms shown in \reffig{fig:perf_compl_2d_scatter} is $0.043$, the values are calculated by determining the absolute difference between SBS and VBS performance for all two-dimensional problems and averaging it. While this suggests there is some potential for algorithm selection, modCMA is the top-performing among the eight algorithms on $6\,665$ out of the $11\,920$ instances ($\approx56\%$). While this indicates that there is some performance complementarity, in order to maximize our ability to differentiate between different algorithm selection settings, we look for subsets of our portfolio which have higher complementarity. To do this systematically, we measure the VBS-SBS gap for each possible subset of our algorithm portfolio. The results of this analysis are shown in \reffig{fig:perf_compl_2d}, where the sets are separated by their corresponding SBS. In this figure, each dot corresponds to one of the items from the powerset, and the shape of the dot encodes the number of solvers in this set. Since the y-axis corresponds to the VBS-SBS gap, the higher the dots, the better the complementarity of the algorithms in the portfolio. By comparing the height of different color dots in each column, we observe that reducing the size of a portfolio by removing a non-SBS solver can only negatively impact the VBS-SBS gap. However, by removing the SBS from a set, we can achieve significant increases in complementarity. The subset with the largest complementarity (triangular dot on the top right of the figure) is a portfolio of 3 algorithms, among which GOMEA performs best. However, since this set contains only three algorithms (RCobyla, GOMEA, and MultiBFGS), we opt to use instead the second-most complementary set, which contains all solvers except for modCMA. With a portfolio of seven algorithms, this still makes for an interesting algorithm selection task. While the same analysis has also been performed for the 5-dimensional functions, suggesting a portfolio of DifferentialEvolution, RCobyla, GOMEA, and MultiBFGS, we decided to use the same portfolio of seven algorithms also in the $d=5$ case. For the latter, the VBS-SBS-gap is $0.083$.

\begin{figure}[h]
  \centering
  \includegraphics{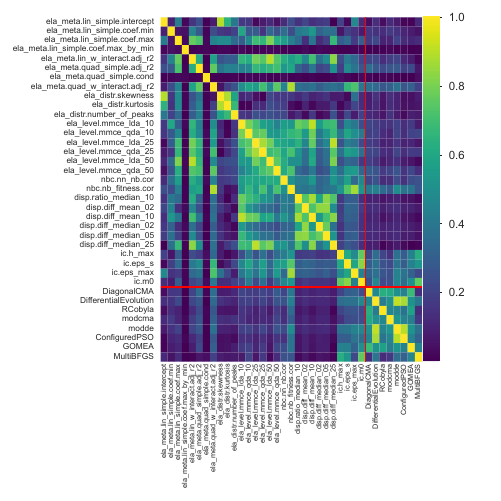}
    \caption{Pearson correlation between all used features and the algorithm performances across all 11\,920 two-dimensional problems. Performances and features are separated by the red lines. No individual feature is able to explain the entire performance of any solver.}
    \label{fig:corr_2d}
\end{figure}

\begin{figure}[h]
  \centering
  \includegraphics{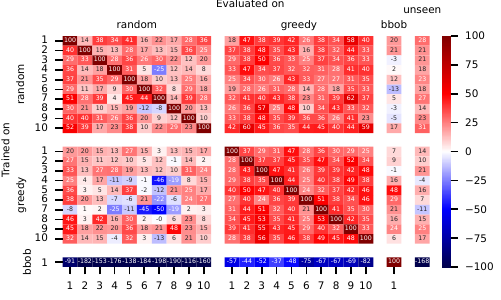}
    \caption{Percentage of average VBS-SBS gap closed on the two-dimensional data subset given by the column, while the AAS model was trained on the two-dimensional data subset given by the row. All training data subsets have size 120 while the evaluation subsets also have size 120 with the exception of the unseen data which has size 11\,800. It is apparent that the sampling, and, thus, distribution of the training data set compared to the distribution of the evaluation data set plays an important role for the model performance.}
    \label{fig:cross_eva_gap_closed_2d}
\end{figure}

\begin{figure}[h]
  \centering
  \includegraphics{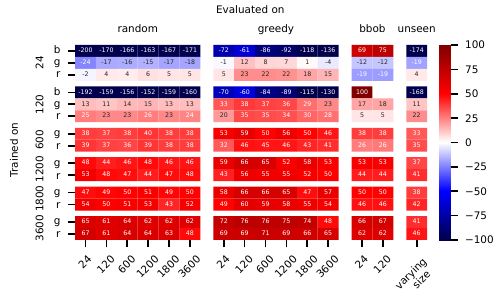}
    \caption{Mean percentage of VBS-SBS gap closed for the two-dimensional case. 
    The performance is averaged over all models trained on all repetitions of subsets given by the rows and evaluated on all evaluation data set repetitions (with the exception of the training data set). 
    Even with rather small training data set sizes, the VBS-SBS gap can be closed to a certain degree, as long as the training set distribution matches the test set distribution. 
    It is possible to mitigate the disadvantage of non-matching training and test data set distributions by increasing the training data set size.}
    \label{fig:avg_diff_sizes}
\end{figure}

\subsection{Automated Algorithm Selection}

With the feature representation and algorithm portfolio established in the previous sections, we can now analyze the relation between these aspects. We start by considering pairwise correlation between all 29 ELA features and AOCC values of all available algorithms, visualized in \reffig{fig:corr_2d}. Here, we can see that while our feature preprocessing removed the most correlated landscape features, there are still some rather large correlations between features within the feature sets. Within the algorithm performance, correlations are generally smaller, which matches our findings from \reffig{fig:perf_compl_2d}. Since the correlation between algorithm performance and individual features is relatively small, it seems unlikely that a univariate model would be sufficient to predict algorithm performance, and we should instead aim to exploit relations between features to build an accurate algorithm selector. 

Our algorithm selection task boils down to predicting the correct algorithm from our portfolio of 7 algorithms, given the 29-dimensional ELA-vector representing the given problem. We opt to use a classification algorithm directly, rather than regression-based selection. The classification model we use is the XGBClassifier~\cite{XGB2016}, with default parameter settings for multi-class classification. These choices are made to keep our pipeline simple~\cite{Kostovska0VDED23ASmodels}.

\paragraph{Impact of sampling strategy}
To analyze the effect of different sampling strategies of training and testing data, we focus on data set sizes of $120$. The reasoning for this choice is twofold -- on one hand this allows us to directly compare to the first five instances of the BBOB functions as one training data set while on the other hand comparable AAS studies often only relied on similar sizes when it comes to training data (e.g. \cite{KerschkeT19} with training data size of 95).
In \reffig{fig:distr_greedy_2d} we show the distribution of problem instances in the 2$d$ feature space that was constructed on the two-dimensional BBOB functions for \reffig{fig:pca_all}. The left column of the figure shows repetitions 1-5 of uniform randomly sampling our set of $11\,920$ problem instances. In the right column we find the first five repetitions of data sets which were sampled using the greedy selection approach (maximizing ELA diversity). All black dots are generated functions and colored dots are BBOB functions, re-using the color bar from the bottom of \reffig{fig:pca_all}.
It is clearly visible that the greedily sampled sets are more evenly spread across the feature space while the uniformly sampled sets resemble the general problem distribution by including more problems in the feature space area with the highest overall problem density. The greedily selected instances, thus, contain a more diverse set of problems which might be expected to benefit the ability of a model to generalize to unseen functions. 
Note that with increasing data set size the greedily selected data sets differ less from the randomly selected ones. To illustrate this effect, we report in \reftab{tab:avg_pairwise_dist} the average pairwise Manhattan distance across all subsets generated with the same data set sizes. 
We see that with increasing data set size, the values for the greedily selected data sets converge to those of the randomly selected data sets. 

\begin{table}
  \caption{Average pairwise Manhattan distance between the functions across all subsets of the indicated data set size.}
  \label{tab:avg_pairwise_dist}
  \begin{tabular}{ccc}
    \toprule
    data set size&greedily selected&randomly selected\\
    \midrule
    24   & 6.81              & 4.22              \\
    120  & 5.78              & 4.35              \\
    600  & 5.18              & 4.51              \\
    1\,200 & 5.04              & 4.62              \\
    1\,800 & 4.82              & 4.55              \\
    3\,600 & 4.60              & 4.53             \\
  \bottomrule
\end{tabular}
\end{table}

In \reffig{fig:cross_eva_gap_closed_2d} we show the performance of all models that were trained on a dataset size of 120 two-dimensional functions. Each row of the heatmaps represents one training scenario as annotated in the figure. The performance is measured in percentage of average SBS-VBS gap closed on the data set given by the column annotation. Blue cells therefore correspond to scenarios where the AAS does not manage to outperform the SBS, whereas the dark red ones are the settings that benefit most from the selector. 

The column "\textbf{unseen}" refers to all two-dimensional problems out of the $11\,920$ that have not been part of the $120$ problems which the respective model was trained on. As visible in the diagonal, all trained models are able to close the average VBS-SBS gap entirely on their respective training data set. This could be a warning sign of over-fitting on the training data. 
With the exception of three scenarios with randomly selected training and test sets, all models manage to outperform the SBS for training sets that were constructed by the same approach.  
We can, therefore, say with some certainty that the models have learned relationships between features and algorithm performances that enable them to make helpful decisions even for unseen data.

Overall, the models trained on greedily selected data sets show a more heterogeneous performance, e.g., the model trained on the greedily selected data set number 7 performs particularly poorly on randomly selected data sets and the unseen data. 
When looking at the performances of models that were trained on either the greedily or randomly selected data and evaluated on sets constructed by the other method, we find that models trained on greedily selected data perform worse on the randomly selected subsets than vice versa. 
This is surprising since, as stated above, our expectations were that models trained on more diverse data would outperform models trained on randomly selected data.
Generally, the randomly selected data sets seem to pose a harder challenge to our learner of choice since even the models trained on randomly selected data sets have more difficulties to perform well on those data sets than on the greedily selected data sets. 

When evaluating both types of models on the first five instances of the BBOB functions, we find that for this scenario, on average, the models trained on greedily selected data come out on top. This could be due to the fact that greedily selected data sets tend to contain more BBOB functions than randomly selected data sets, as partly visible in \reffig{fig:distr_greedy_2d}.
If we instead consider the performances on all unseen data per model, models trained on randomly selected data sets are the better choice.
These results are an indication that not the feature-space diversity but rather matching distributions of training and testing data sets are the decisive factor that drive the performance on the test sets.

Finally, looking at the model trained on the first five instances of the BBOB functions, we see that it is not able to outperform the SBS in any of the evaluation scenarios -- confirming our belief that training AAS models only on BBOB does not generalize.

\paragraph{Impact of training data set size}
Lastly, we look at the effect of different training data set sizes on the performance of the models. 
This is a particularly interesting analysis since it is expensive to evaluate new solvers on large sets of benchmark functions to re-train an AAS model. 
\reffig{fig:avg_diff_sizes} again shows the percentage of average VBS-SBS gap closed, but this time the performances are aggregated by taking the mean over all repetitions of data sets with a certain size both as training and evaluation data sets. The ticks on the y-axis with labels 'b', 'g' and 'r' represent the different data selection methods 'bbob', 'greedy', and 'random'. To bring this into context with \reffig{fig:cross_eva_gap_closed_2d} and make it a little bit easier to understand -- the value in \reffig{fig:avg_diff_sizes} that is declared as trained on randomly selected data sets with size $120$ and evaluated on randomly selected data sets with size $120$ corresponds to the average of all values shown in the top left heat map in \reffig{fig:cross_eva_gap_closed_2d}, putting aside the value of the diagonal (we ignore this scenario because it corresponds to identical training and evaluation data). 

We observe that a training data set size of $24$ is not sufficient to achieve satisfying performances of the AAS models. Still, there already are observable differences in performance between the differently selected training data sets. A random selection is favorable for all evaluation scenarios with the exception of the BBOB functions. However, since using the BBOB functions as training set is primed to be better and the models trained on greedily selected functions are also not able to improve upon the SBS performance, this scenario does not carry much weight. 

Throughout all training data set sizes, we find that models, trained on a data set with matching sampling strategy to the respective evaluation data set, are able to close the VBS-SBS gap further than models trained on data with a different sampling strategy.
Values that are consistently slightly smaller within one evaluation and training category correspond to evaluation instances where the data set sizes and selection strategies of evaluation and training data sets match. In these cases we have entirely disjoint training and evaluation data sets. Thus, for the models trained on randomly selected data sets these values are similar to the performances these models achieve on the unseen data sets since both subsets share the same distribution. For models trained on greedily selected data sets these specific values are also consistently a little lower, but still higher than their performance on unseen data because greedily selected evaluation data sets do not share a distribution with the unseen data. 
For greedily trained models we lose performance with increasing evaluation data set size since the distribution becomes closer and closer to the overall distribution and, thus, starts to differ from the training set distribution.
Within one training data set size the models trained on randomly selected data always outperform the other models when evaluating on unseen data. We take this as another sign for the importance of training data set distribution for AAS. 
We also find that when increasing the training set size to the next higher level, the performance on unseen data of models trained on the greedily selected data catches up to the performance of models that were trained on randomly sampled data with a smaller training size. Thus, by increasing the size of the training data set we can mitigate the effects of non-matching training and evaluation data distributions.
The effect of non-matching distribution becomes stronger, the smaller the training data set size is. For example, models that were trained on greedily sampled data sets need at least a training data set size of 120 to achieve any gap closure on unseen data. This is not the case for models trained on randomly selected data sets which have a matching distribution with the unseen data. These models are able to achieve a mean gap closure of $4\%$ with a training data set size of 24.
For models trained on the BBOB functions, \reffig{fig:avg_diff_sizes} confirms that they are unable to generalize well.

\section{Conclusions}\label{sec:conclusion}
Using the MA-BBOB function suite as an environment, we have analyzed in this work the impact of training set selection on the generalization ability of feature-based algorithm selectors. As has possibly been expected, selectors trained on instances whose distribution matches the one of the testing sets tend to perform much better than those trained on instances that follow different distributions. The results highlight an important bottleneck for the application of automated algorithm selection techniques in practical optimization scenarios: unlike the settings typically considered in the Machine Learning literature~\cite{HutterKV19}, in our applications we often lack knowledge of the distribution of the problem instances that we are faced with at execution time. It is not even uncommon that we are asked to help optimizing a single problem instance. We are therefore interested in algorithm selection models that are trained to make good recommendations across broad ranges of possible applications, without a priori knowledge of their nature. 

 We also observed (cf. Fig.~\ref{fig:avg_diff_sizes}) that the negative effects of non-matching distributions of training and test instances can be mitigated by increasing the size of the training sets. However, the more training instances we consider, the more computational effort is required to gather the 
 data needed for training the selector. This trade-off between accuracy and consumption of resources should be carefully balanced in future development of algorithm selection approaches, especially when the goal is to develop toolboxes that can be extended by adding more algorithms or new feature sets.

 We considered three different training instance selection methods in this work. Different ways to select instances are well conceivable, e.g., making use of algorithm performances, of the weights used to generate the MA-BBOB functions, or using hybridization between these options. In addition, we do not need to limit ourselves to data sets for which algorithm performance data is homogeneously available for all training instances. In fact, it is very well possible that heterogeneous training sets, possibly even non-overlapping ones, are favorable for training algorithm selection methods. 
We conjecture such approaches to be particularly interesting when coupled with multi-target regression or classification approaches.  
The algorithm-specific instances could be wisely selected, e.g., using algorithm footprint techniques as proposed in~\cite{footprintsGECCO2023}, or -- this may be the more relevant case for practical applications -- simply using whatever data is already available. 

We consider the MA-BBOB problem generator a meaningful environment to obtain first insights into the above-mentioned questions. However, we are also aware that the real challenge in practice is the generalization ability beyond ``academic'' benchmarking suites. 

\textbf{Acknowledgments.} This work was realized with the financial support of ANR project ANR-22-ERCS-0003-01 and of CNRS Sciences informatiques project \emph{IOHprofiler}.
\bibliography{refs} 


\begin{thebibliography}{41}


\ifx \showCODEN    \undefined \def \showCODEN     #1{\unskip}     \fi
\ifx \showDOI      \undefined \def \showDOI       #1{#1}\fi
\ifx \showISBNx    \undefined \def \showISBNx     #1{\unskip}     \fi
\ifx \showISBNxiii \undefined \def \showISBNxiii  #1{\unskip}     \fi
\ifx \showISSN     \undefined \def \showISSN      #1{\unskip}     \fi
\ifx \showLCCN     \undefined \def \showLCCN      #1{\unskip}     \fi
\ifx \shownote     \undefined \def \shownote      #1{#1}          \fi
\ifx \showarticletitle \undefined \def \showarticletitle #1{#1}   \fi
\ifx \showURL      \undefined \def \showURL       {\relax}        \fi
\providecommand\bibfield[2]{#2}
\providecommand\bibinfo[2]{#2}
\providecommand\natexlab[1]{#1}
\providecommand\showeprint[2][]{arXiv:#2}

\bibitem[B{\"a}ck et~al\mbox{.}(2023)]%
        {back2023evolutionary}
\bibfield{author}{\bibinfo{person}{Thomas H.~W. B{\"a}ck}, \bibinfo{person}{Anna~V. Kononova}, \bibinfo{person}{Bas van Stein}, \bibinfo{person}{Hao Wang}, \bibinfo{person}{Kirill~A. Antonov}, \bibinfo{person}{Roman~T. Kalkreuth}, \bibinfo{person}{Jacob de Nobel}, \bibinfo{person}{Diederick Vermetten}, \bibinfo{person}{Roy de Winter}, {and} \bibinfo{person}{Furong Ye}.} \bibinfo{year}{2023}\natexlab{}.
\newblock \showarticletitle{Evolutionary Algorithms for Parameter Optimization—Thirty Years Later}.
\newblock \bibinfo{journal}{\emph{Evolutionary Computation}} \bibinfo{volume}{31}, \bibinfo{number}{2} (\bibinfo{year}{2023}), \bibinfo{pages}{81~--~122}.
\newblock


\bibitem[Bartz{-}Beielstein et~al\mbox{.}(2020)]%
        {TBB20benchmarking}
\bibfield{author}{\bibinfo{person}{Thomas Bartz{-}Beielstein}, \bibinfo{person}{Carola Doerr}, \bibinfo{person}{Jakob Bossek}, \bibinfo{person}{Sowmya Chandrasekaran}, \bibinfo{person}{Tome Eftimov}, \bibinfo{person}{Andreas Fischbach}, \bibinfo{person}{Pascal Kerschke}, \bibinfo{person}{Manuel L{\'{o}}pez{-}Ib{\'{a}}{\~{n}}ez}, \bibinfo{person}{Katherine~M. Malan}, \bibinfo{person}{Jason~H. Moore}, \bibinfo{person}{Boris Naujoks}, \bibinfo{person}{Patryk Orzechowski}, \bibinfo{person}{Vanessa Volz}, \bibinfo{person}{Markus Wagner}, {and} \bibinfo{person}{Thomas Weise}.} \bibinfo{year}{2020}\natexlab{}.
\newblock \showarticletitle{Benchmarking in Optimization: Best Practice and Open Issues}.
\newblock \bibinfo{journal}{\emph{CoRR}}  \bibinfo{volume}{abs/2007.03488} (\bibinfo{year}{2020}).
\newblock
\showeprint[arxiv]{2007.03488}
\urldef\tempurl%
\url{https://arxiv.org/abs/2007.03488}
\showURL{%
\tempurl}


\bibitem[Belkhir et~al\mbox{.}(2017)]%
        {BelkhirDSS17}
\bibfield{author}{\bibinfo{person}{Nacim Belkhir}, \bibinfo{person}{Johann Dr{\'{e}}o}, \bibinfo{person}{Pierre Sav{\'{e}}ant}, {and} \bibinfo{person}{Marc Schoenauer}.} \bibinfo{year}{2017}\natexlab{}.
\newblock \showarticletitle{Per instance algorithm configuration of {CMA-ES} with limited budget}. In \bibinfo{booktitle}{\emph{Proceedings of Genetic and Evolutionary Computation Conference (GECCO)}}. \bibinfo{publisher}{ACM}, \bibinfo{pages}{681--688}.
\newblock
\urldef\tempurl%
\url{https://doi.org/10.1145/3071178.3071343}
\showDOI{\tempurl}


\bibitem[Cenikj et~al\mbox{.}(2022)]%
        {GECCO2022SELECTOR}
\bibfield{author}{\bibinfo{person}{Gjorgjina Cenikj}, \bibinfo{person}{Ryan~Dieter Lang}, \bibinfo{person}{Andries~Petrus Engelbrecht}, \bibinfo{person}{Carola Doerr}, \bibinfo{person}{Peter Korošec}, {and} \bibinfo{person}{Tome Eftimov}.} \bibinfo{year}{2022}\natexlab{}.
\newblock \showarticletitle{SELECTOR: Selecting a Representative Benchmark Suite for Reproducible Statistical Comparison}. In \bibinfo{booktitle}{\emph{Proceedings of Genetic and Evolutionary Computation Conference (GECCO)}}. \bibinfo{publisher}{ACM}.
\newblock
\urldef\tempurl%
\url{https://doi.org/10.1145/3512290.3528809}
\showDOI{\tempurl}


\bibitem[Chen and Guestrin(2016)]%
        {XGB2016}
\bibfield{author}{\bibinfo{person}{Tianqi Chen} {and} \bibinfo{person}{Carlos Guestrin}.} \bibinfo{year}{2016}\natexlab{}.
\newblock \showarticletitle{{XGBoost}: A Scalable Tree Boosting System}. In \bibinfo{booktitle}{\emph{Proceedings of the 22nd ACM SIGKDD International Conference on Knowledge Discovery and Data Mining}} \emph{(\bibinfo{series}{KDD '16})}. \bibinfo{publisher}{ACM}, \bibinfo{address}{New York, NY, USA}, \bibinfo{pages}{785--794}.
\newblock
\showISBNx{978-1-4503-4232-2}
\urldef\tempurl%
\url{https://doi.org/10.1145/2939672.2939785}
\showDOI{\tempurl}


\bibitem[de~Nobel et~al\mbox{.}(2023)]%
        {iohexperimenter}
\bibfield{author}{\bibinfo{person}{Jacob de Nobel}, \bibinfo{person}{Furong Ye}, \bibinfo{person}{Diederick Vermetten}, \bibinfo{person}{Hao Wang}, \bibinfo{person}{Carola Doerr}, {and} \bibinfo{person}{Thomas B{\"a}ck}.} \bibinfo{year}{2023}\natexlab{}.
\newblock \showarticletitle{Iohexperimenter: Benchmarking platform for iterative optimization heuristics}.
\newblock \bibinfo{journal}{\emph{Evolutionary Computation}} (\bibinfo{year}{2023}), \bibinfo{pages}{1--6}.
\newblock


\bibitem[Dietrich and Mersmann(2022)]%
        {DietrichM22affinebbob}
\bibfield{author}{\bibinfo{person}{Konstantin Dietrich} {and} \bibinfo{person}{Olaf Mersmann}.} \bibinfo{year}{2022}\natexlab{}.
\newblock \showarticletitle{Increasing the Diversity of Benchmark Function Sets Through Affine Recombination}. In \bibinfo{booktitle}{\emph{Proceedings of Parallel Problem Solving from Nature (PPSN'22)}} \emph{(\bibinfo{series}{LNCS}, Vol.~\bibinfo{volume}{13398})}, \bibfield{editor}{\bibinfo{person}{G{\"{u}}nter Rudolph}, \bibinfo{person}{Anna~V. Kononova}, \bibinfo{person}{Hern{\'{a}}n~E. Aguirre}, \bibinfo{person}{Pascal Kerschke}, \bibinfo{person}{Gabriela Ochoa}, {and} \bibinfo{person}{Tea Tusar}} (Eds.). \bibinfo{publisher}{Springer}, \bibinfo{pages}{590--602}.
\newblock
\urldef\tempurl%
\url{https://doi.org/10.1007/978-3-031-14714-2\_41}
\showDOI{\tempurl}


\bibitem[Dietrich et~al\mbox{.}(2024)]%
        {reproducibility_and_figures}
\bibfield{author}{\bibinfo{person}{Konstantin Dietrich}, \bibinfo{person}{Diederick Vermetten}, \bibinfo{person}{Carola Doerr}, {and} \bibinfo{person}{Pascal Kerschke}.} \bibinfo{year}{2024}\natexlab{}.
\newblock \bibinfo{title}{Reproducibility files and additional figures}.
\newblock \bibinfo{howpublished}{Code and data repository (Zenodo): \url{doi.org/10.5281/zenodo.10911264} Figure repository (Figshare): \url{doi.org/10.6084/m9.figshare.25127501}}.
\newblock


\bibitem[Hansen et~al\mbox{.}(2021)]%
        {hansen2021coco}
\bibfield{author}{\bibinfo{person}{Nikolaus Hansen}, \bibinfo{person}{Anne Auger}, \bibinfo{person}{Raymond Ros}, \bibinfo{person}{Olaf Mersmann}, \bibinfo{person}{Tea Tu{\v{s}}ar}, {and} \bibinfo{person}{Dimo Brockhoff}.} \bibinfo{year}{2021}\natexlab{}.
\newblock \showarticletitle{COCO: A platform for comparing continuous optimizers in a black-box setting}.
\newblock \bibinfo{journal}{\emph{Optimization Methods and Software}} \bibinfo{volume}{36}, \bibinfo{number}{1} (\bibinfo{year}{2021}), \bibinfo{pages}{114--144}.
\newblock
\urldef\tempurl%
\url{https://doi.org/10.1080/10556788.2020.1808977}
\showURL{%
\tempurl}


\bibitem[Hansen et~al\mbox{.}(2009)]%
        {bbobfunctions}
\bibfield{author}{\bibinfo{person}{Nikolaus Hansen}, \bibinfo{person}{Steffen Finck}, \bibinfo{person}{Raymond Ros}, {and} \bibinfo{person}{Anne Auger}.} \bibinfo{year}{2009}\natexlab{}.
\newblock \bibinfo{booktitle}{\emph{{Real-Parameter Black-Box Optimization Benchmarking 2009: Noiseless Functions Definitions}}}.
\newblock \bibinfo{type}{{T}echnical {R}eport} RR-6829. \bibinfo{institution}{{INRIA}}.
\newblock
\urldef\tempurl%
\url{https://hal.inria.fr/inria-00362633/document}
\showURL{%
\tempurl}


\bibitem[Hutter et~al\mbox{.}(2019)]%
        {HutterKV19}
\bibfield{editor}{\bibinfo{person}{Frank Hutter}, \bibinfo{person}{Lars Kotthoff}, {and} \bibinfo{person}{Joaquin Vanschoren}} (Eds.). \bibinfo{year}{2019}\natexlab{}.
\newblock \bibinfo{booktitle}{\emph{Automated Machine Learning - Methods, Systems, Challenges}}.
\newblock \bibinfo{publisher}{Springer}.
\newblock
\showISBNx{978-3-030-05317-8}
\urldef\tempurl%
\url{https://doi.org/10.1007/978-3-030-05318-5}
\showDOI{\tempurl}


\bibitem[Kerschke et~al\mbox{.}(2019)]%
        {KerschkeHNT19}
\bibfield{author}{\bibinfo{person}{Pascal Kerschke}, \bibinfo{person}{Holger~H. Hoos}, \bibinfo{person}{Frank Neumann}, {and} \bibinfo{person}{Heike Trautmann}.} \bibinfo{year}{2019}\natexlab{}.
\newblock \showarticletitle{Automated Algorithm Selection: Survey and Perspectives}.
\newblock \bibinfo{journal}{\emph{Evolutionary Computation}} \bibinfo{volume}{27}, \bibinfo{number}{1} (\bibinfo{year}{2019}), \bibinfo{pages}{3--45}.
\newblock
\urldef\tempurl%
\url{https://doi.org/10.1162/evco\_a\_00242}
\showDOI{\tempurl}


\bibitem[Kerschke et~al\mbox{.}(2015)]%
        {Kerschke2015}
\bibfield{author}{\bibinfo{person}{Pascal Kerschke}, \bibinfo{person}{Mike Preuss}, \bibinfo{person}{Simon Wessing}, {and} \bibinfo{person}{Heike Trautmann}.} \bibinfo{year}{2015}\natexlab{}.
\newblock \showarticletitle{Detecting funnel structures by means of exploratory landscape analysis}.
\newblock \bibinfo{journal}{\emph{GECCO 2015 - Proceedings of the 2015 Genetic and Evolutionary Computation Conference}} (\bibinfo{year}{2015}), \bibinfo{pages}{265--272}.
\newblock
\showISBNx{9781450334723}
\urldef\tempurl%
\url{https://doi.org/10.1145/2739480.2754642}
\showDOI{\tempurl}


\bibitem[Kerschke and Trautmann(2019a)]%
        {KerschkeT19}
\bibfield{author}{\bibinfo{person}{Pascal Kerschke} {and} \bibinfo{person}{Heike Trautmann}.} \bibinfo{year}{2019}\natexlab{a}.
\newblock \showarticletitle{Automated Algorithm Selection on Continuous Black-Box Problems by Combining Exploratory Landscape Analysis and Machine Learning}.
\newblock \bibinfo{journal}{\emph{Evolutionary Computation}} \bibinfo{volume}{27}, \bibinfo{number}{1} (\bibinfo{year}{2019}), \bibinfo{pages}{99--127}.
\newblock
\urldef\tempurl%
\url{https://doi.org/10.1162/evco\_a\_00236}
\showDOI{\tempurl}


\bibitem[Kerschke and Trautmann(2019b)]%
        {Kerschke2019flacco}
\bibfield{author}{\bibinfo{person}{Pascal Kerschke} {and} \bibinfo{person}{Heike Trautmann}.} \bibinfo{year}{2019}\natexlab{b}.
\newblock \showarticletitle{Comprehensive Feature-Based Landscape Analysis of Continuous and Constrained Optimization Problems Using the R-package flacco}.
\newblock In \bibinfo{booktitle}{\emph{Applications in Statistical Computing -- From Music Data Analysis to Industrial Quality Improvement}}, \bibfield{editor}{\bibinfo{person}{Nadja Bauer}, \bibinfo{person}{Katja Ickstadt}, \bibinfo{person}{Karsten Lübke}, \bibinfo{person}{Gero Szepannek}, \bibinfo{person}{Heike Trautmann}, {and} \bibinfo{person}{Maurizio Vichi}} (Eds.). Vol.~\bibinfo{volume}{17}. \bibinfo{publisher}{Springer}, \bibinfo{pages}{93--123}.
\newblock
\urldef\tempurl%
\url{https://doi.org/10.1007/978-3-030-25147-5_7}
\showDOI{\tempurl}


\bibitem[Kostovska et~al\mbox{.}(2022)]%
        {kostovska2022per}
\bibfield{author}{\bibinfo{person}{Ana Kostovska}, \bibinfo{person}{Anja Jankovic}, \bibinfo{person}{Diederick Vermetten}, \bibinfo{person}{Jacob de Nobel}, \bibinfo{person}{Hao Wang}, \bibinfo{person}{Tome Eftimov}, {and} \bibinfo{person}{Carola Doerr}.} \bibinfo{year}{2022}\natexlab{}.
\newblock \showarticletitle{Per-run algorithm selection with warm-starting using trajectory-based features}. In \bibinfo{booktitle}{\emph{International Conference on Parallel Problem Solving from Nature}}. Springer, \bibinfo{pages}{46--60}.
\newblock


\bibitem[Kostovska et~al\mbox{.}(2023)]%
        {Kostovska0VDED23ASmodels}
\bibfield{author}{\bibinfo{person}{Ana Kostovska}, \bibinfo{person}{Anja Jankovic}, \bibinfo{person}{Diederick Vermetten}, \bibinfo{person}{Saso Dzeroski}, \bibinfo{person}{Tome Eftimov}, {and} \bibinfo{person}{Carola Doerr}.} \bibinfo{year}{2023}\natexlab{}.
\newblock \showarticletitle{Comparing Algorithm Selection Approaches on Black-Box Optimization Problems}. In \bibinfo{booktitle}{\emph{Proceedings of Genetic and Evolutionary Computation Conference (GECCO, companion material)}}. \bibinfo{publisher}{{ACM}}, \bibinfo{pages}{495--498}.
\newblock
\urldef\tempurl%
\url{https://doi.org/10.1145/3583133.3590697}
\showDOI{\tempurl}


\bibitem[Long et~al\mbox{.}(2023a)]%
        {long2023challenges}
\bibfield{author}{\bibinfo{person}{Fu~Xing Long}, \bibinfo{person}{Diederick Vermetten}, \bibinfo{person}{Anna~V Kononova}, \bibinfo{person}{Roman Kalkreuth}, \bibinfo{person}{Kaifeng Yang}, \bibinfo{person}{Thomas B{\"a}ck}, {and} \bibinfo{person}{Niki van Stein}.} \bibinfo{year}{2023}\natexlab{a}.
\newblock \showarticletitle{Challenges of ELA-guided Function Evolution using Genetic Programming}.
\newblock \bibinfo{journal}{\emph{arXiv preprint arXiv:2305.15245}} (\bibinfo{year}{2023}).
\newblock


\bibitem[Long et~al\mbox{.}(2023b)]%
        {long2023bbob}
\bibfield{author}{\bibinfo{person}{Fu~Xing Long}, \bibinfo{person}{Diederick Vermetten}, \bibinfo{person}{Bas van Stein}, {and} \bibinfo{person}{Anna~V. Kononova}.} \bibinfo{year}{2023}\natexlab{b}.
\newblock \showarticletitle{BBOB Instance Analysis: Landscape Properties and Algorithm Performance Across Problem Instances}. \bibinfo{publisher}{Springer-Verlag}, \bibinfo{pages}{380–395}.
\newblock
\showISBNx{978-3-031-30228-2}
\urldef\tempurl%
\url{https://doi.org/10.1007/978-3-031-30229-9_25}
\showDOI{\tempurl}


\bibitem[Lunacek and Whitley(2006)]%
        {Lunacek2006}
\bibfield{author}{\bibinfo{person}{Monte Lunacek} {and} \bibinfo{person}{Darrell Whitley}.} \bibinfo{year}{2006}\natexlab{}.
\newblock \showarticletitle{The dispersion metric and the CMA evolution strategy}.
\newblock \bibinfo{journal}{\emph{GECCO 2006 - Genetic and Evolutionary Computation Conference}}  \bibinfo{volume}{1} (\bibinfo{year}{2006}), \bibinfo{pages}{477--484}.
\newblock
\showISBNx{1595931864}
\urldef\tempurl%
\url{https://doi.org/10.1145/1143997.1144085}
\showDOI{\tempurl}


\bibitem[López-Ibáñez et~al\mbox{.}(2024)]%
        {eafecdf}
\bibfield{author}{\bibinfo{person}{Manuel López-Ibáñez}, \bibinfo{person}{Diederick Vermetten}, \bibinfo{person}{Johann Dreo}, {and} \bibinfo{person}{Carola Doerr}.} \bibinfo{year}{2024}\natexlab{}.
\newblock \bibinfo{title}{Using the Empirical Attainment Function for Analyzing Single-objective Black-box Optimization Algorithms}.
\newblock
\newblock
\showeprint[arxiv]{2404.02031}


\bibitem[Mersmann et~al\mbox{.}(2011)]%
        {mersmann2011exploratory}
\bibfield{author}{\bibinfo{person}{Olaf Mersmann}, \bibinfo{person}{Bernd Bischl}, \bibinfo{person}{Heike Trautmann}, \bibinfo{person}{Mike Preuss}, \bibinfo{person}{Claus Weihs}, {and} \bibinfo{person}{G{\"u}nter Rudolph}.} \bibinfo{year}{2011}\natexlab{}.
\newblock \showarticletitle{Exploratory landscape analysis}. In \bibinfo{booktitle}{\emph{Proceedings of Genetic and Evolutionary Computation Conference (GECCO)}}. \bibinfo{publisher}{ACM}, \bibinfo{pages}{829--836}.
\newblock
\urldef\tempurl%
\url{https://doi.org/10.1145/2001576.2001690}
\showDOI{\tempurl}


\bibitem[Meunier et~al\mbox{.}(2020)]%
        {MeunierDRT20}
\bibfield{author}{\bibinfo{person}{Laurent Meunier}, \bibinfo{person}{Carola Doerr}, \bibinfo{person}{J{\'{e}}r{\'{e}}my Rapin}, {and} \bibinfo{person}{Olivier Teytaud}.} \bibinfo{year}{2020}\natexlab{}.
\newblock \showarticletitle{Variance Reduction for Better Sampling in Continuous Domains}. In \bibinfo{booktitle}{\emph{Proceedings of Parallel Problem Solving from Nature (PPSN)}} \emph{(\bibinfo{series}{LNCS}, Vol.~\bibinfo{volume}{12269})}. \bibinfo{publisher}{Springer}, \bibinfo{pages}{154--168}.
\newblock
\urldef\tempurl%
\url{https://doi.org/10.1007/978-3-030-58112-1\_11}
\showDOI{\tempurl}


\bibitem[Morgan and Gallagher(2014)]%
        {MorganG14}
\bibfield{author}{\bibinfo{person}{Rachael Morgan} {and} \bibinfo{person}{Marcus Gallagher}.} \bibinfo{year}{2014}\natexlab{}.
\newblock \showarticletitle{Sampling Techniques and Distance Metrics in High Dimensional Continuous Landscape Analysis: Limitations and Improvements}.
\newblock \bibinfo{journal}{\emph{{IEEE} Trans. Evol. Comput.}} \bibinfo{volume}{18}, \bibinfo{number}{3} (\bibinfo{year}{2014}), \bibinfo{pages}{456--461}.
\newblock
\urldef\tempurl%
\url{https://doi.org/10.1109/TEVC.2013.2281521}
\showDOI{\tempurl}


\bibitem[Mu\~{n}oz et~al\mbox{.}(2015)]%
        {Munoz2015}
\bibfield{author}{\bibinfo{person}{Mario~Andr\'{e}s Mu\~{n}oz}, \bibinfo{person}{Michael Kirley}, {and} \bibinfo{person}{Saman~K. Halgamuge}.} \bibinfo{year}{2015}\natexlab{}.
\newblock \showarticletitle{Exploratory landscape analysis of continuous space optimization problems using information content}.
\newblock \bibinfo{journal}{\emph{IEEE Transactions on Evolutionary Computation}}  \bibinfo{volume}{19} (\bibinfo{year}{2015}), \bibinfo{pages}{74--87}.
\newblock
Issue 1.
\showISSN{1089778X}
\urldef\tempurl%
\url{https://doi.org/10.1109/TEVC.2014.2302006}
\showDOI{\tempurl}


\bibitem[Mu{\~{n}}oz and Smith{-}Miles(2020)]%
        {MunozS20spacefillingBBO}
\bibfield{author}{\bibinfo{person}{Mario~A. Mu{\~{n}}oz} {and} \bibinfo{person}{Kate Smith{-}Miles}.} \bibinfo{year}{2020}\natexlab{}.
\newblock \showarticletitle{Generating New Space-Filling Test Instances for Continuous Black-Box Optimization}.
\newblock \bibinfo{journal}{\emph{Evol. Comput.}} \bibinfo{volume}{28}, \bibinfo{number}{3} (\bibinfo{year}{2020}), \bibinfo{pages}{379--404}.
\newblock
\urldef\tempurl%
\url{https://doi.org/10.1162/evco\_a\_00262}
\showDOI{\tempurl}


\bibitem[Nikolikj et~al\mbox{.}(2023a)]%
        {nikolikj2023rf}
\bibfield{author}{\bibinfo{person}{Ana Nikolikj}, \bibinfo{person}{Carola Doerr}, {and} \bibinfo{person}{Tome Eftimov}.} \bibinfo{year}{2023}\natexlab{a}.
\newblock \showarticletitle{RF+ clust for Leave-One-Problem-Out Performance Prediction}. In \bibinfo{booktitle}{\emph{International Conference on the Applications of Evolutionary Computation (Part of EvoStar)}}. Springer, \bibinfo{pages}{285--301}.
\newblock


\bibitem[Nikolikj et~al\mbox{.}(2023b)]%
        {footprintsGECCO2023}
\bibfield{author}{\bibinfo{person}{Ana Nikolikj}, \bibinfo{person}{Saso Dzeroski}, \bibinfo{person}{Mario~Andr{\'{e}}s Mu{\~{n}}oz}, \bibinfo{person}{Carola Doerr}, \bibinfo{person}{Peter Korosec}, {and} \bibinfo{person}{Tome Eftimov}.} \bibinfo{year}{2023}\natexlab{b}.
\newblock \showarticletitle{Algorithm Instance Footprint: Separating Easily Solvable and Challenging Problem Instances}. In \bibinfo{booktitle}{\emph{Proceedings of Genetic and Evolutionary Computation Conference (GECCO)}}. \bibinfo{publisher}{{ACM}}, \bibinfo{pages}{529--537}.
\newblock
\urldef\tempurl%
\url{https://doi.org/10.1145/3583131.3590424}
\showDOI{\tempurl}


\bibitem[Prager and Trautmann(2023)]%
        {Pflacco2023}
\bibfield{author}{\bibinfo{person}{Raphael~Patrick Prager} {and} \bibinfo{person}{Heike Trautmann}.} \bibinfo{year}{2023}\natexlab{}.
\newblock \showarticletitle{{Pflacco: Feature-Based Landscape Analysis of Continuous and Constrained Optimization Problems in Python}}.
\newblock \bibinfo{journal}{\emph{Evolutionary Computation}} (\bibinfo{date}{07} \bibinfo{year}{2023}), \bibinfo{pages}{1--25}.
\newblock
\showISSN{1063-6560}
\urldef\tempurl%
\url{https://doi.org/10.1162/evco_a_00341}
\showDOI{\tempurl}


\bibitem[Rapin and Teytaud(2018)]%
        {nevergrad}
\bibfield{author}{\bibinfo{person}{J{\'{e}}r{\'{e}}my Rapin} {and} \bibinfo{person}{Olivier Teytaud}.} \bibinfo{year}{2018}\natexlab{}.
\newblock \bibinfo{title}{{Nevergrad - A gradient-free optimization platform}}.
\newblock \bibinfo{howpublished}{\url{https://GitHub.com/FacebookResearch/Nevergrad}}.
\newblock


\bibitem[Renau et~al\mbox{.}(2021)]%
        {RenauDDD21evoapps}
\bibfield{author}{\bibinfo{person}{Quentin Renau}, \bibinfo{person}{Johann Dr{\'{e}}o}, \bibinfo{person}{Carola Doerr}, {and} \bibinfo{person}{Benjamin Doerr}.} \bibinfo{year}{2021}\natexlab{}.
\newblock \showarticletitle{Towards Explainable Exploratory Landscape Analysis: Extreme Feature Selection for Classifying {BBOB} Functions}. In \bibinfo{booktitle}{\emph{Proceedings of Applications of Evolutionary Computation (EvoApplications)}} \emph{(\bibinfo{series}{LNCS}, Vol.~\bibinfo{volume}{12694})}. \bibinfo{publisher}{Springer}, \bibinfo{pages}{601--617}.
\newblock
\urldef\tempurl%
\url{https://doi.org/10.1007/978-3-030-72699-7\_2}
\showDOI{\tempurl}


\bibitem[Rice(1976)]%
        {Rice76}
\bibfield{author}{\bibinfo{person}{John~R. Rice}.} \bibinfo{year}{1976}\natexlab{}.
\newblock \showarticletitle{The Algorithm Selection Problem}.
\newblock \bibinfo{journal}{\emph{Advances in Computers}}  \bibinfo{volume}{15} (\bibinfo{year}{1976}), \bibinfo{pages}{65--118}.
\newblock
\urldef\tempurl%
\url{https://doi.org/10.1016/S0065-2458(08)60520-3}
\showDOI{\tempurl}


\bibitem[Seiler et~al\mbox{.}(2022)]%
        {SeilerPKT22featurefree}
\bibfield{author}{\bibinfo{person}{Moritz~Vinzent Seiler}, \bibinfo{person}{Raphael~Patrick Prager}, \bibinfo{person}{Pascal Kerschke}, {and} \bibinfo{person}{Heike Trautmann}.} \bibinfo{year}{2022}\natexlab{}.
\newblock \showarticletitle{A collection of deep learning-based feature-free approaches for characterizing single-objective continuous fitness landscapes}. In \bibinfo{booktitle}{\emph{Proceedings of Genetic and Evolutionary Computation Conference (GECCO)}}. \bibinfo{publisher}{{ACM}}, \bibinfo{pages}{657--665}.
\newblock
\urldef\tempurl%
\url{https://doi.org/10.1145/3512290.3528834}
\showDOI{\tempurl}


\bibitem[{\v{S}}kvorc et~al\mbox{.}(2020)]%
        {vskvorc2020understanding}
\bibfield{author}{\bibinfo{person}{Urban {\v{S}}kvorc}, \bibinfo{person}{Tome Eftimov}, {and} \bibinfo{person}{Peter Koro{\v{s}}ec}.} \bibinfo{year}{2020}\natexlab{}.
\newblock \showarticletitle{Understanding the problem space in single-objective numerical optimization using exploratory landscape analysis}.
\newblock \bibinfo{journal}{\emph{Applied Soft Computing}}  \bibinfo{volume}{90} (\bibinfo{year}{2020}), \bibinfo{pages}{106138}.
\newblock
\showISSN{1568-4946}
\urldef\tempurl%
\url{https://doi.org/10.1016/j.asoc.2020.106138}
\showDOI{\tempurl}


\bibitem[{\v{S}}kvorc et~al\mbox{.}(2022)]%
        {vskvorc2022transfer}
\bibfield{author}{\bibinfo{person}{Urban {\v{S}}kvorc}, \bibinfo{person}{Tome Eftimov}, {and} \bibinfo{person}{Peter Koro{\v{s}}ec}.} \bibinfo{year}{2022}\natexlab{}.
\newblock \showarticletitle{Transfer Learning Analysis of Multi-Class Classification for Landscape-Aware Algorithm Selection}.
\newblock \bibinfo{journal}{\emph{Mathematics}} \bibinfo{volume}{10}, \bibinfo{number}{3} (\bibinfo{year}{2022}).
\newblock
\showISSN{2227-7390}
\urldef\tempurl%
\url{https://doi.org/10.3390/math10030432}
\showDOI{\tempurl}


\bibitem[Tian et~al\mbox{.}(2020)]%
        {tian2020arecommender}
\bibfield{author}{\bibinfo{person}{Ye Tian}, \bibinfo{person}{Shichen Peng}, \bibinfo{person}{Xingyi Zhang}, \bibinfo{person}{Tobias Rodemann}, \bibinfo{person}{Kay~Chen Tan}, {and} \bibinfo{person}{Yaochu Jin}.} \bibinfo{year}{2020}\natexlab{}.
\newblock \showarticletitle{A Recommender System for Metaheuristic Algorithms for Continuous Optimization Based on Deep Recurrent Neural Networks}.
\newblock \bibinfo{journal}{\emph{IEEE Transactions on Artificial Intelligence}} \bibinfo{volume}{1}, \bibinfo{number}{1} (\bibinfo{year}{2020}), \bibinfo{pages}{5--18}.
\newblock
\urldef\tempurl%
\url{https://doi.org/10.1109/TAI.2020.3022339}
\showDOI{\tempurl}


\bibitem[van Rijn et~al\mbox{.}(2016)]%
        {modular-CMAES}
\bibfield{author}{\bibinfo{person}{Sander van Rijn}, \bibinfo{person}{Hao Wang}, \bibinfo{person}{Matthijs van Leeuwen}, {and} \bibinfo{person}{Thomas B{\"{a}}ck}.} \bibinfo{year}{2016}\natexlab{}.
\newblock \showarticletitle{Evolving the structure of Evolution Strategies}. In \bibinfo{booktitle}{\emph{Proceedings of {IEEE} Symposium Series on Computational Intelligence (SSCI)}}. \bibinfo{publisher}{IEEE}, \bibinfo{pages}{1--8}.
\newblock
\urldef\tempurl%
\url{https://doi.org/10.1109/SSCI.2016.7850138}
\showDOI{\tempurl}


\bibitem[Vermetten et~al\mbox{.}(2023a)]%
        {modde}
\bibfield{author}{\bibinfo{person}{Diederick Vermetten}, \bibinfo{person}{Fabio Caraffini}, \bibinfo{person}{Anna~V. Kononova}, {and} \bibinfo{person}{Thomas B{\"{a}}ck}.} \bibinfo{year}{2023}\natexlab{a}.
\newblock \showarticletitle{Modular Differential Evolution}. In \bibinfo{booktitle}{\emph{Proceedings of the Genetic and Evolutionary Computation Conference, {GECCO} 2023, Lisbon, Portugal, July 15-19, 2023}}, \bibfield{editor}{\bibinfo{person}{Sara Silva} {and} \bibinfo{person}{Lu{\'{\i}}s Paquete}} (Eds.). \bibinfo{publisher}{{ACM}}, \bibinfo{pages}{864--872}.
\newblock
\urldef\tempurl%
\url{https://doi.org/10.1145/3583131.3590417}
\showDOI{\tempurl}


\bibitem[Vermetten et~al\mbox{.}(2023c)]%
        {mabbob_arxiv}
\bibfield{author}{\bibinfo{person}{Diederick Vermetten}, \bibinfo{person}{Furong Ye}, \bibinfo{person}{Thomas B{\"a}ck}, {and} \bibinfo{person}{Carola Doerr}.} \bibinfo{year}{2023}\natexlab{c}.
\newblock \showarticletitle{MA-BBOB: A Problem Generator for Black-Box Optimization Using Affine Combinations and Shifts}.
\newblock \bibinfo{journal}{\emph{arXiv preprint arXiv:2312.11083}} (\bibinfo{year}{2023}).
\newblock


\bibitem[Vermetten et~al\mbox{.}(2023d)]%
        {MABBOBautoML}
\bibfield{author}{\bibinfo{person}{Diederick Vermetten}, \bibinfo{person}{Furong Ye}, \bibinfo{person}{Thomas Bäck}, {and} \bibinfo{person}{Carola Doerr}.} \bibinfo{year}{2023}\natexlab{d}.
\newblock \showarticletitle{{MA-BBOB:} Many-Affine Combinations of {BBOB} Functions for Evaluating {AutoML} Approaches in Noiseless Numerical Black-Box Optimization Contexts}. In \bibinfo{booktitle}{\emph{Proceedings of the AutoML Conference (AutoML)}}. \bibinfo{publisher}{{PMLR}}.
\newblock
\newblock
\shownote{Available at \url{https://openreview.net/forum?id=uN70Dum6pC2}}.


\bibitem[Vermetten et~al\mbox{.}(2023b)]%
        {ABBOB-GECCO}
\bibfield{author}{\bibinfo{person}{Diederick Vermetten}, \bibinfo{person}{Furong Ye}, {and} \bibinfo{person}{Carola Doerr}.} \bibinfo{year}{2023}\natexlab{b}.
\newblock \showarticletitle{Using Affine Combinations of {BBOB} Problems for Performance Assessment}. In \bibinfo{booktitle}{\emph{Proceedings of Genetic and Evolutionary Computation Conference (GECCO'23)}}, Vol.~\bibinfo{volume}{abs/2303.04573}. \bibinfo{publisher}{ACM}.
\newblock
\urldef\tempurl%
\url{https://doi.org/10.1145/3583131.3590412}
\showDOI{\tempurl}


\end{thebibliography}
\bibliographystyle{ACM-Reference-Format}
\end{document}